\documentclass{article}
\usepackage{ijcai23}

\usepackage{times}
\usepackage{soul}
\usepackage{url}
\usepackage[hidelinks]{hyperref}
\usepackage[utf8]{inputenc}
\usepackage[small]{caption}
\usepackage{graphicx}
\usepackage{amsmath}
\usepackage{amsthm}
\usepackage{booktabs}
\usepackage{algorithm}
\usepackage{algorithmic}
\usepackage[switch]{lineno}
\usepackage{multirow}
\usepackage{bbding}
\usepackage{tikz}
\usepackage{amssymb}

\title{Guided Patch-Grouping Wavelet Transformer with Spatial Congruence for Ultra-High Resolution Segmentation}

\author{
Deyi Ji$^{1,2}$
\and
Feng Zhao$^1$\thanks{Corresponding Author.}\and
Hongtao Lu$^{3,4}$
\affiliations
$^1$University of Science and Technology of China\\
$^2$Alibaba Group\\
$^3$Department of Computer Science and Engineering,  Shanghai Jiao Tong University\\
$^4$MOE Key Lab of Artificial Intelligence, AI Institute, Shanghai Jiao Tong University
\emails
jideyi@mail.ustc.edu.cn,
fzhao956@ustc.edu.cn,
htlu@sjtu.edu.cn
}

\begin{document}

\maketitle

\begin{abstract}

Most existing ultra-high resolution (UHR) segmentation methods always struggle in the dilemma of balancing memory cost and local characterization accuracy, which are both taken into account in our proposed Guided Patch-Grouping Wavelet Transformer (GPWFormer) that achieves impressive performances.  In this work, GPWFormer is a Transformer ($\mathcal{T}$)-CNN ($\mathcal{C}$) mutual leaning framework, where $\mathcal{T}$ takes the whole UHR image as input and harvests both local details and fine-grained long-range contextual dependencies, while $\mathcal{C}$ takes downsampled image as input for learning the category-wise deep context. For the sake of high inference speed and low computation complexity, 
$\mathcal{T}$ partitions the original UHR image into patches and groups them dynamically, then learns the low-level local details with the  lightweight multi-head Wavelet Transformer (WFormer) network.
Meanwhile, the fine-grained long-range contextual dependencies are also captured during this process, since patches that are far away in the spatial domain can also be assigned to the same group. In addition, masks produced by $\mathcal{C}$ are utilized to guide the patch grouping process, providing a heuristics decision. Moreover, the congruence constraints between the two branches are also exploited to maintain the spatial consistency among the patches. Overall, we stack the multi-stage process in a pyramid way. Experiments show that GPWFormer  outperforms the existing methods with significant improvements on five benchmark datasets.

\end{abstract}

\section{Introduction}
\label{intro}

The analysis of ultra-high resolution (UHR) geospatial image with millions or even billions of pixels has opened new horizons for the computer vision community, playing an increasingly important role in a wide range of geosciences and urban construction applications, such as disaster control, environmental monitoring, land resource protection and  urban planning \cite{glnet,isdnet,urur}. The focus of this paper is on semantic segmentation, providing a better understanding by assigning each pixel into a specified category. 

Fully convolution neural networks (FCN) based methods have driven rapid growth in segmentation for regular resolution images, but overlook the feasibility of larger scale input. Due to the memory limitation, earliest works for UHR image segmentation basically follow two paradigms: (1) downsampling the image to a regular resolution, or (2) cropping the image into small patches,  feeding them to network sequentially and  merging their predictions. Intuitively both the two paradigms will result in inaccurate results, the former loses many local details while the latter lacks of global context. After that, methods specially designed for UHR images are proposed and most of them follow the global-local collaborative framework to preserve both global and local information with two deep branches, taking the downsampled entire image and cropped local patches as inputs respectively. The most representative works are GLNet \cite{glnet} and FCtL \cite{fctl}. Despite the considerable performance, their memory cost is high and inference speed is very low, due to the deep branches and sequentially inference. Later, following the bilateral architecture \cite{bisenet}, ISDNet \cite{isdnet} proposes to combine a shallow and a deep branch. The shallow branch takes the whole UHR image as input and extracts multi-scale shallow features, while the deep branch takes the highly downsampled image as input to extract one deep feature, then the features are fused for final prediction with a specially designed fusion module. This type of architecture avoids local patches cropping and sequentially prediction thus improves the inference speed with a large margin, but also results in weaker performance, especially in local characterization, since the local details will never be fully addressed in the shallow branch with UHR input. In addition, their proposed fusion module also introduces much extra memory cost. In a word, due to enormous pixels, existing methods for UHR segmentation usually struggle in the pressing dilemma of balancing memory cost and local characterization accuracy. 

To this end, in this paper we propose a novel Guided Patch-Grouping Wavelet Transformer (GPWFormer) network to address the above balance problem for UHR image segmentation. In general, we formulate a hybrid CNN-Transformer framework in dual-branch style,
where the Transformer branch takes locally cropped UHR image as input and harvests both local details and fine-grained long-range dependencies, while the CNN branch takes downsampled UHR image as input for learning the category-wise deep context. For the sake of high inference speed and low computation complexity, different from the classical global-local framework, the local patches are fed into the Transformer branch all at once and dynamically grouped, then each group is fed into a different Transformer head to extract local texture details. Meanwhile fine-grained long-range dependencies are also captured during this process, since patches that are far away in spatial domain can also be assigned to a same group. In addition, inspired by \cite{wavevit},  invertible downsampling operations with dense wavelets are integrated into the Transformer for lightweight memory cost, and masks produced by CNN branch are utilized to guide the patch grouping process, providing a heuristics decision. Moreover, the congruence constraint between the two branches are also exploited to maintain the spatial consistency among the patches. In general, we stack the multi-stage process in a pyramid way.

Overall, our contributions are summarized as follows:

\begin{itemize}
    \item We propose a novel Patch-Grouping Wavelet Transformer (GPWFormer) network for ultra-high resolution image segmentation, which is a hybrid CNN-Transformer in dual-branch style to harvest fine-grained both low-level and high-level context  simultaneously in an efficient way. 

    \item Specifically, we introduce a Wavelet Transformer to integrate the Transformer branch with dense wavelets for lightweight memory cost, and further decrease its computation complexity with heuristics grouping masks guided by the corresponding deep features of  CNN branch.
    
    \item Moreover, the congruence constraints between the two branches are also exploited to maintain the spatial consistency among the patches.
    
    \item Extensive experiments demonstrate that the proposed GPWFormer obtains an excellent balance between memory cost and local segmentation accuracy, and outperforms the existing methods with significant improvements on five public datasets.
\end{itemize}

\section{Related Work}

\subsection{Generic Semantic Segmentation}
Deep learning based methods have taken a big step forward on computer vision \cite{goodfellow2016deep,resnet,ji2019end,feng2018challenges,ipgn,wang2021learning,cagcn}. The development of deep CNN and Transformer over the past few years has driven rapid growth of methods in generic  semantic segmentation for natural images and daily photos \cite{cdgc,deeplabv3+,stlnet}.
Earlier semantic segmentation models for generic images was mainly based on the fully convolutional networks (FCN) \cite{fcn} and recent ones followed the design of Transformer network. FCN-based methods usually relied on large receptive field and fine-grained deep features, such as  DeepLab \cite{deeplabv2,deeplabv3+}, DANet \cite{danet}, OCRNet \cite{ocrnet}. While Transformer-based networks viewed segmentation  as a Sequence-to-Sequence perspective and have become a new research hotspot. Representative works included SETR \cite{SETR} and Swin \cite{swin}. However, Transformer networks usually took amounts of memory cost and computation complexity, limiting its development on UHR image segmentation. In this paper, we aim to take advantage of the strong representation ability of Transformer meanwhile decrease its memory cost for UHR segmentation.
In addition, knowledge distillation methods \cite{sstkd} have also been applied to make the network lightweight.

\subsection{Ultra-High Resolution Image Segmentation}
Benefited from the advancement of photography and sensor technologies, the accessibility and analysis of ultra-high resolution geospatial images has opened new horizons for the computer vision community, playing an increasingly important role in a wide range of geosciences and urban construction applications, including but not limited to disaster control, environmental monitoring, land resource protection and  urban planning \cite{urur}. According to \cite{ascher2007filmmaker,glnet}, an image with at least $4.1\times 10^6$ pixels can reach the minimum bar of ultra-high definition media, usually deriving from a wide range of scientific applications, for example, geospatial and histopathological images. For ultra-high resolution image segmentation, CascadePSP \cite{cascadepsp} proposed to improve the coarse segmentation results with a pretrained model to generate high-quality results.
GLNet \cite{glnet}  incorporated both global and local information deeply in a two-stream branch manner. 
FCtL \cite{fctl} exploited a squeeze-and-split structure to fuse multi-scale features information. 
ISDNet \cite{isdnet} integrated the shallow and deep networks. These existing works lack of a further in-depth analysis and have obvious drawbacks analyzed in the Introduction section. WSDNet \cite{urur} was proposed as an efficient and effective framework for UHR segmentation especially with ultra-rich context.

\begin{figure*}[ht]
    \includegraphics[width=0.98\linewidth]{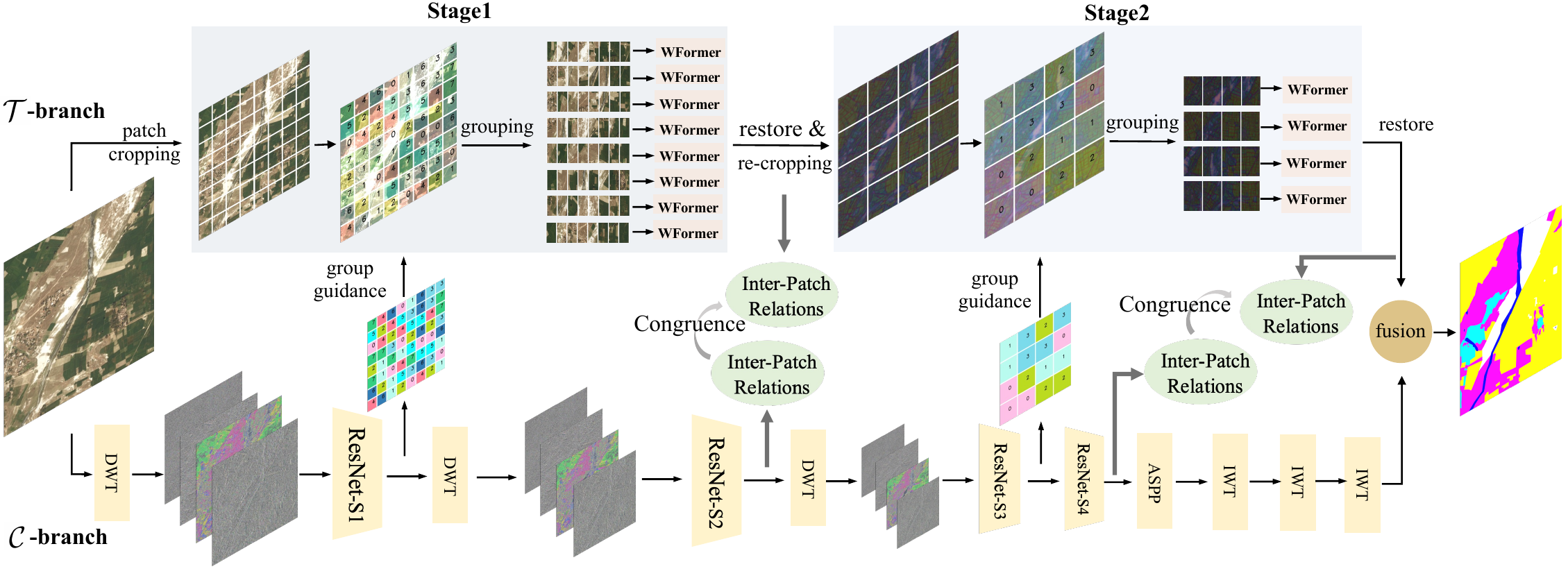}
    \caption{The overview of the proposed GPWFormer. Upper: Transformer branch ($\mathcal{T}$); Lower: CNN branch ($\mathcal{C}$). In each stage of $\mathcal{T}$, the input UHR image/feature is evenly partitioned into patches and fed into the network all at once. Then, they are grouped guided by a mask generated from the corresponding features in $\mathcal{C}$. Next, intra-group relations in each group are learned with their respective WFormer heads. In $\mathcal{C}$, DeeplabV3+ (ResNet18) integrated with pyramid wavelets is employed, taking downsampled UHR image with wavelet transform as input to capture deep category-wise context (``ResNet-S1" means the first stage of ResNet and so on). After each stage of $\mathcal{T}$, we maintain the inter-patch relation consistency with a congruence constraint by $\mathcal{C}$. Finally, the outputs of $\mathcal{T}$ and $\mathcal{C}$ are fused and supervised with a focal loss.}
    \label{method}
\end{figure*}

\section{Method}

In this section, we introduce the proposed Guided Patch-Grouping Wavelet Transformer (GPWFormer) in detail. Firstly, we introduce the overall structure. Subsequently, we illustrate the Guided Patch-Grouping strategy and Wavelet Transformer respectively. Next the details of spatial congruence are explained. 
\subsection{Overview}
\label{overview}

Figure \ref{method} shows the overall framework of the proposed GPWFormer, which consists of dual branches. The upper branch is a lightweight Transformer network taking the original UHR image as input to harvest both local structured details and fine-grained long-range spatial dependencies, while the lower is a deep CNN network taking the downsampled UHR image as input to learn category-wise deep context. For simplicity, we denote the two branches as $\mathcal{T}$ and $\mathcal{C}$ respectively. In $\mathcal{C}$, any classical generic segmentation architecture can be utilized, here  DeepLabV3+ with a lightweight backbone ResNet18 is employed. Besides, in order to further reduce its computation complexity, we integrate each stage of ResNet18 with discrete wavelet transform (DWT) to reduce the dimension of intermediate features, followed by the 
Atrous Spatial Pyramid Pooling (ASPP) module and inverse wavelet transform (IWT).

In $\mathcal{T}$, the input UHR image is firstly evenly partition into local patches. Different from  existing global-local frameworks that take into patches sequentially, all the patches are fed into $\mathcal{T}$ at once in our framework, which can greatly accelerate the inference process. In order to decrease the memory cost in this situation, we divide these patches into multiple groups, then intra-group relations for each group are learned with a corresponding shallow Wavelet Transformer (WFormer) head respectively, which is proposed for 
more effective and efficient learning. The relations consist of both local structured relations and long-range dependencies, since both spatially adjacent and nonadjacent patches can be assigned to a same group. Meanwhile all the low-level pixel-wise details are also captured during the process. Multiple stages can be stacked in a pyramid way for fully characterization, and the stage number is set to 2 here for higher inference speed, which can also be set to a flexible number according to the practical needs. The number of groups among stages are set in a pyramid manner to obtain representations of different granularity. It is worthy noted that  the corresponding features from $\mathcal{C}$ are employed to guide the patch-grouping, providing a  heuristics decision. Moreover, after the learning of each stage in $\mathcal{T}$, we also add a congruence constraint to the inter-patch relations by the corresponding stage of $\mathcal{C}$ to maintain the spatial consistency. Finally the output of $\mathcal{T}$ and $\mathcal{C}$ are fused and supervised with a focal loss. 

\subsection{Guided Patch-Grouping} 

In each stage of $\mathcal{T}$, we produce a guidance for patch-grouping from the corresponding feature from $\mathcal{C}$, as shown in Figure \ref{method}, and the first stage is taken  as example for the following illustration. Given the UHR input $I\in \mathbb{R}^{H_{I} \times W_{I}}$ in $\mathcal{T}$, we evenly partition it into $m \times n$ patches, which are then divided into $G$ groups for the subsequent WFormer heads. Let $h, w$ denote the \textit{height} and \textit{width} of each patch respectively. Our aim is to produce a guidance mask $M \in \mathbb{R}^{m \times n}$ from the $\mathcal{C}$, $M_{i,j} \in [1, G]$ denotes the group index of patch ${(i, j)}$, where $(i \in [1, m], j \in [1, n])$.

Concretely, we employ the low-frequency subband $F^{LL} \in \mathbb{R}^{C_1 \times H_1 \times W_1}$ of wavelet transform for the feature after \textit{ResNet-S1}  in $\mathcal{C}$ to produce $M$, since low-frequency subband is able to preserve more spatial details. $C_1, H_1$, and $W_1$ denote the \textit{channel}, \textit{height}, \textit{width} of $F_{LL}$ respectively. Firstly, channel of $F_{LL}$ is decreased to $G$ with an intuitive PCA method instead of a typical convolution operator, so that the whole mask-producing process is not necessary to be learnable, thereby reducing the amount of calculation, while avoiding the design of complex forward propagation and gradient backward processes. Next following the same partition operations for $I$ in $\mathcal{T}$,  we apply a patch-wise average pooling to $F^{LL}$, resulting in the mask feature $F^{M} \in \mathbb{R}^{G \times m \times n}$. Formally, the process of patch-wise average pooling is denoted as,
\begin{equation}
\begin{aligned}
    F^M_{i, j} &= GAP(F^{LL}_{\{i \times h_1\ : \ (i+1) \times h_1, \ j \times w_1 \ : \ (j+1) \times w_1\}}) \\
    h_1 &= H_1 / m, w_1 = W_1 / n, i \in [1, m], j \in [1, n],
\end{aligned}
\label{pooling}
\end{equation}

\begin{figure}[ht]
    \centering
    \includegraphics[width=1\linewidth]{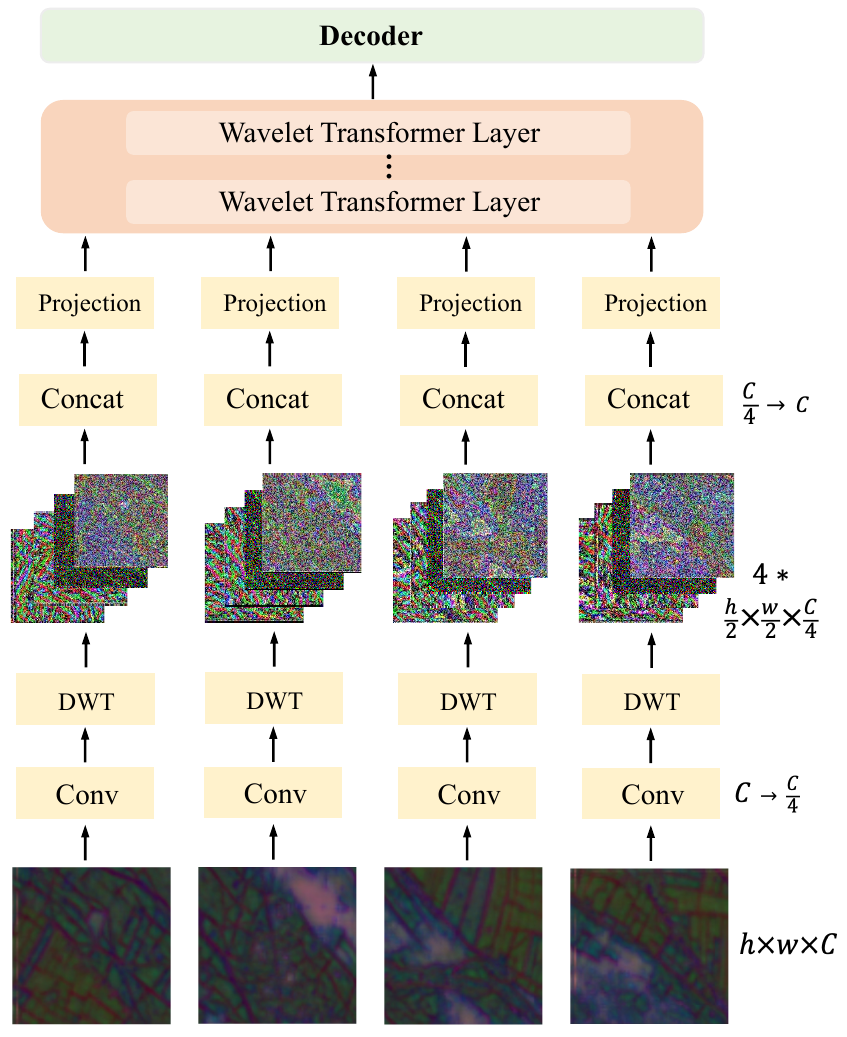}
    \caption{The details of WFormer.}
    \label{wsa}
\end{figure}

\noindent where $GAP(\cdot)$ denotes global average pooling. The above process can also be viewed as an average pooling with \textit{row\_stride} $=h_1$ and \textit{column\_stride} $=w_1$. Then we apply a $softmax$ function to $F^M$, generate the score mask $S \in \mathbb{R}^{G \times m \times n}$, and argmax $S$ along the last two dimension to generate the mask $M$, 
\begin{equation}
    M = \mathop{\arg\max}_{m,n} (S).
\end{equation}

Noted that $M$ may not divide the groups evenly, that is,  not all groups may have equal number of patches, and some groups may have more than $\frac{m\times n}{G}$ patches, thus we sort the patches in descending order of scores in these groups respectively, then re-distribute their last extra patches to other groups based on scores.

\subsection{Wavelet Transformer}

In our framework, the input to each Transformer head is a group of patch features. Taking any group in first stage of $\mathcal{T}$ as example for illustration, let $\{F_{u} \in \mathbb{R}^{h \times w \times C} \}$ denote the set of input patch features in the group, where $u$ is patch index. Following the design of \cite{wavevit}, we introduce the Wavelet Transformer to make the network lightweight.  For each $F_{u}$, we first reduce its channel to $C / 4$ with a convolution layer, then 2D DWT is utilized to decompose its four wavelet subbands $\{F_{u}^{LL}, F_{u}^{LH}, F_{u}^{LL}, F_{u}^{LL}\}$ with four filters $f_{LL}, f_{LH}, f_{HL}, f_{HH}$. The dimension of both the subbands is $\mathbb{R}^{\frac{h}{2} \times \frac{w}{2} \times \frac{C}{4}}$. $F_{u}^{LL}$ is the low-frequency subbands while the others are high-frequency ones, next the four subbands are concatenated to $\breve{F}_{u} \in \mathbb{R}^{\frac{h}{2} \times \frac{w}{2} \times C}$. By this way, the dimension of all patch features are reduced so that the computation complexity is squarely decreased. Noted that despite the downsampling operation is deployed, due to the biorthogonal property of DWT, the original feature can be accurately reconstructed by the inverse wavelet transform (IWT) in subsequent steps. Finally, $\{\breve{F}_{u}\}$ is projected into a sequence $\breve{F} \in \mathbb{R}^{L\times C}$ and embedded spatial information as previous general Transformer networks \cite{SETR}, and  formulated as input to corresponding Transformer head. 

As shown in Figure \ref{wsa}, following the general design of  Transformer for  segmentation, WFormer also  contains an encoder and a decoder, where encoder consists of $L_e$ layers of multi-head self-attention (MSA) and MLP blocks. The input to self-attention of each layer $l$ is in a triplet of ($Q, K, V$), calculated by the current input $Y^{l-1} \in \mathbb{R}^{L\times C}$ (the input for the first layer is $\breve{F}$), with three learn-able linear projection layers $\mathbf{W}_Q, \mathbf{W}_{K}, \mathbf{W}_{V} \in \mathbb{R}^{C\times d}$ ($d$ is the dimension), as:
\begin{equation}
    {Q = Y^{l-1}\mathbf{W}_Q, K = Y^{l-1}\mathbf{W}_K, V = Y^{l-1}\mathbf{W}_V} .
\end{equation}

Noted that the computational complexity of ordinary Self-Attention (SA) grows quadratically with dimensions, so we further apply the similar procedure of wavelet transform to $K$ and $V$ to decrease the size, denoted as $\breve{K}$ and $\breve{V}$, by which we propose the Wavelet-SA (WSA) and Wavelet-MSA (WMSA), formulated as:
\begin{equation}
    WSA(Y^{l-1}) = Y^{l-1} +  softmax(\frac{Q\breve{K}^{\top}}{\sqrt{d}}) (\breve{V}),
\end{equation}
\begin{equation}
    \resizebox{0.9\hsize}{!}{$WMSA(Y^{l-1}) = [SA_1(Y^{l-1}), SA_2(Y^{l-1}),... , SA_m(Y^{l-1})]\mathbf{W}_O$},
\end{equation}
\noindent where $m$ is number of WSA operations
in WMSA, $\mathbf{W}_O$ is the transformation matrix. Then IWT is utilized to restore the same dimension of output as the input. In the light of basic theory of DWT and IWT, WFormer is able to harvest stronger both local details and spatial long-range dependencies without loss of essential information, compared to Transformer with traditional pooling operations. 

\subsection{Spatial Consistency Congruence}

With multiple Transformer heads, the patches of input feature are learned in groups, which affects the translation-invariant and results in the spatial in-consistency among the patches. Since $\mathcal{C}$ learns the input image globally and is able to preserve the overall spatial consistency, we propose to utilize  congruence constraints from $\mathcal{C}$ to each stage of $\mathcal{T}$, to maintain the spatial congruence of inter-patch relations in $\mathcal{T}$. 

Considering the stage $s$ in $\mathcal{T}$, given the patch features set $\{P^{\mathcal{T}_s}_k\}$ $(k \in [1, m_s\times n_s])$ after WFormer heads, where $m_s \times n_s$ is the number of patches in stage $s$ of $\mathcal{T}$ (denoted as $\mathcal{T}_s$), we first calculate their inter-patch relations $C^{\mathcal{T}}_s$, which defines as the sum of relations between any two patch centers. The center of a patch is defined as the average of its all pixel features. To capture the high-order relations in an efficient manner, inspired by the theory of SVM \cite{rbf}, we employ the metric with Gaussian Radial Based Function (RBF) kernel to calculate the inter-patch relations. Therefore $C^{\mathcal{T}}_s$ can be formulated as,
\begin{equation}
    C^{\mathcal{T}}_s = \sum_{k_1=1}^{m_s \times n_s}\sum_{k_2=1}^{m_s \times n_s} R(A(P^{\mathcal{T}_s}_{k_1}), A(P^{\mathcal{T}_s}_{k_2})),
\end{equation}
\noindent where $R(\cdot), A(\cdot)$ are the RBF kernel and Average function respectively. To simplify the calculation, we replace the RBF calculation with its sum of $T$-order Taylor series,
\begin{equation}
\begin{aligned}
    R(A(P^{\mathcal{T}_s}_{k_1}), & A(P^{\mathcal{T}_s}_{k_2})) = e^{-\theta \cdot || A(P^{\mathcal{T}_s}_{k_1}) - A(P^{\mathcal{T}_s}_{k_2})||^2} \\
    & = \sum_{t=0}^{T}  \frac{(2\theta)^t}{t!}(A(P^{\mathcal{T}_s}_{k_1}) \cdot A(P^{\mathcal{T}_s}_{k_2})^{\top})^t e^{-2\theta},
\end{aligned}
\end{equation}
\noindent where $\theta$ is temperature parameter, $\top$ is matrix transposition operator. 

Similarly, given the feature in corresponding stage of $\mathcal{C}$, we partition it as the same style of $\{P^{\mathcal{T}_s}_k\}$, and calculate its inter-patch relations $C^{\mathcal{C}}_s$, so the spatial consistency constraint can be formulated,
\begin{equation}
\begin{aligned}
    C &= \sum_{s} \frac{1}{m_s \times n_s}||C^{\mathcal{C}}_s - C^{\mathcal{T}}_s||^2_2,
\end{aligned}
\end{equation}
\noindent which is implemented as a loss cooperating with the main focal loss.

So the overall loss $\mathcal{L}$ is a weighted combination of a main focal loss and the above spatial consistency constraint,
\begin{equation}
    \mathcal{L} = \mathcal{L}_{focal} + \alpha C,
\end{equation}
\noindent where $\alpha$ is the loss weight and set to 0.8.

\section{Experiments}

\subsection{Datasets and Evaluation Metrics}
In order to validate the effectiveness of our proposed method in a wide perspective, we perform experiments on five datasets, including DeepGlobe, Inria Aerial, Cityscapes, ISIC and CRAG.

\subsubsection{DeepGlobe}

The DeepGlobe dataset \cite{deepglobe} have 803 UHR images, with 455/207/142 images for training, validation and testing. Each image contains $2448 \times 2448$ pixels and the dense annotation contains seven classes of landscape regions.

\subsubsection{Inria Aerial}

The Inria Aerial Challenge dataset \cite{inria} has 180 UHR images captured from five cities. Each image contains $5000 \times 5000$ pixels and is annotated with a binary mask for building/non-building areas, with 126/27/27 images for training, validation and testing.

\begin{table}[ht]
    \centering
    \scalebox{1}{\begin{tabular}{c c | c c c | c c}
    \toprule
    CNN & 
    \begin{tabular}[c]{@{}c@{}} Trans \\ former \end{tabular} & 
    P.G. &
    Wave. &
    Cong. &
    \begin{tabular}[c]{@{}c@{}} mIoU \\ (\%) \end{tabular} & \begin{tabular}[c]{@{}c@{}} Mem \\ (M) \end{tabular} \\ \midrule
    \checkmark &  &  &  &  & 62.7 & - \\ 
    & \checkmark & \checkmark & \checkmark & \checkmark & 73.6 & - \\ \midrule
    \checkmark & \checkmark & \checkmark & \checkmark & \checkmark & 75.8 & 2380 \\ 
    \checkmark & \checkmark & \checkmark & \checkmark &  & 74.5 & 2380 \\ 
    \checkmark & \checkmark & \checkmark &  & \checkmark & 75.9 & 3370 \\ 
    \checkmark & \checkmark &  & \checkmark &  & 76.1 & 6090 \\

    \bottomrule
    \end{tabular}}
    \caption{Effectiveness of each component in GPWFormer. ``P.G., Wave., Cong." indicate Patch-Grouping, Wavelet Transform, and Spatial Congruence, respectively.}
    \label{exp_GPWFormer}
\end{table}

\begin{table}[ht]
    \centering
    \scalebox{1}{\begin{tabular}{c | c c c c  c}
    \toprule
    \begin{tabular}[c]{@{}c@{}} Grouping \\ Strategy \end{tabular} & 
    \begin{tabular}[c]{@{}c@{}} Linear \\ (row) \end{tabular} & 
    \begin{tabular}[c]{@{}c@{}} Linear \\ (column) \end{tabular} & 
    Rectangle  & Guided \\ \midrule
    mIoU & 74.1 & 74.0 & 74.8 &  75.8  \\ 
    \bottomrule
    \end{tabular}}
    \label{exp:GPWFormer}
    \caption{Comparisons of patch-grouping strategy.}
\end{table}

\begin{table}[ht]
    \centering
    \scalebox{1}{\begin{tabular}{c | c c c c c  }
    \toprule
    Metric & Mean & Inner-Dot &
    \begin{tabular}[c]{@{}c@{}} RBF \\ order-1 \end{tabular} &
    \begin{tabular}[c]{@{}c@{}} RBF \\ order-2 \end{tabular} &
    \begin{tabular}[c]{@{}c@{}} RBF \\ order-3 \end{tabular}  \\ \midrule
    mIoU & 74.7 & 74.7 & 75.0 & 75.5 & 75.8 \\ 
    \bottomrule
    \end{tabular}}
    \caption{The impact of spatial congruence.}
    \label{exp_congruence}
\end{table}

\begin{table}[ht]
\centering
\scalebox{1}{\begin{tabular}{c|ccc}
\toprule
Method                   & 
\begin{tabular}[c]{@{}c@{}} Average \\ Pooling \end{tabular} &  
\begin{tabular}[c]{@{}c@{}} Transposed \\ Convolution \end{tabular} & 
\begin{tabular}[c]{@{}c@{}} Wavelet \\ Transform \end{tabular} \\ \midrule
\ mIoU                 &  74.8  & 75.0   &  75.8     \\
Mem                      &  2219  & 2701   &  2380     \\ \bottomrule
\end{tabular}}
\caption{Comparisons of downsampling methods.}
\label{exp_downsample}
\end{table}

\subsubsection{CityScapes}

The Cityscapes dataset \cite{cityscapes} has 5,000 images of 19 semantic classes, with 2,979/500/1,525 images for training, validation and testing.

\subsubsection{ISIC}

The ISIC Lesion Boundary Segmentation Challenge dataset \cite{isic} contains 2596 UHR images, with 2077/260/259 images for training, validation and testing.

\subsubsection{CRAG}

CRAG \cite{crag} dataset includes two classes and exhibits different differentiated glandular morphology, with 173/40 images for training/testing. Their average size is 1512 $\times$ 1516.

\subsubsection{Evaluation Metrics}

In all experiments, we adopt the mIoU, F1 score, Accuracy and memory cost to study the effectiveness. 

\subsection{Implementation Details}

We use the MMSegmentation codebase \cite{mmsegmentation} following the default augmentations without bells and whistles, and train on a server with Tesla V100 GPUs with batch size 8. During the image cropping of Transformer branch, the size of patches are $500 \times 500$ pixels and neighboring patches have $120 \times 580$ pixels overlap region in order to avoid boundary vanishing. The input image of CNN branch is also downsampled to size $500 \times 500$ to trade-off performance and efficiency. We employ three layers in WFormer encoder and  pre-train the Transformer on the Imagenet-1K dataset. During training, Focal loss \cite{focal} with $\lambda=2$ is used to supervise training, and Adam \cite{adam} optimizer is used in the optimization process. We set the initial learning rate to $5 \times 10^{-5}$, and it is decayed by a poly learning rate policy where the initial learning rate is multiplied by $(1 - \frac{iter}{total\_iter})^{0.9}$ after each iteration. The maximum iteration number set to 40k, 80k, 160k and 80k for Inria Aerial, DeepGlobe, Cityscapes and ISIC respectively. We use the command line tool “gpustat” to measure the GPU memory, with the mini-batch size of 1 and avoid calculating any gradients.

\subsection{Ablation Study}

In this section, we delve into the modules and settings of our proposed model and demonstrate their effectiveness. All the ablation studies are performed on DeepGlobe \textit{test} set.

\subsubsection{Effectiveness of GPWFormer} 

We conduct experiments to verify the effectiveness of different components in GPWFormer, as shown in Table \ref{exp_GPWFormer}. Firstly, only CNN branch or Transformer branch will lead to lightweight memory cost but lower performance, indicating the effectiveness of the hybrid CNN-Transformer architecture. Then dual-branch with all components achieves impressive and balanced results. Removing ``Spatial Congruence" module leads to a lower mIoU since the spatial consistency among groups is mismatched, while memory cost is steady as this module is only used in training. Using an ordinary Transformer without wavelets has a less memory, showing that WFormer is able to improve the efficiency while maintaining a comparable performance. Applying WFormer on the whole image results in better performance while severely degraded inference speed, proving that patch-grouping strategy indeed decreases the computation complexity.

\subsubsection{Comparisons of Patch-Grouping Strategies} 

We show the superiority of the proposed guided patch-grouping strategy in Table \ref{exp_congruence}, and  several other strategies are also utilized for comparison. ``Linear (row)" and ``Linear (column)" mean grouping the patches in order of rows and columns respectively, and ``Rectangle" means in the order of rectangular boxes. Experiments show that guided grouping shows best performance, proving that CNN branch indeed provides an effective guidance for patch grouping of Transformer branch.

\begin{table}[ht]
    \centering
    \scalebox{1}{\begin{tabular}{l c c c c }
      \toprule
      \textbf{Generic Model} & 
      \begin{tabular}[c]{@{}c@{}} mIoU \\ (\%)$\uparrow$ \end{tabular} & 
      \begin{tabular}[c]{@{}c@{}} F1 \\ (\%)$\uparrow$ \end{tabular} &  \begin{tabular}[c]{@{}c@{}} Acc \\ (\%)$\uparrow$ \end{tabular} & \begin{tabular}[c]{@{}c@{}} Mem \\ (M)$\downarrow$ \end{tabular}  
      \\ \midrule
    \textbf{\textit{Local Inference}} & & & &   \\
    U-Net & 37.3  & -   & -  &  949  \\
    DeepLabv3+ & 63.1 & -  & - & 1279    \\
    FCN-8s & 71.8  & 82.6 & 87.6 & 1963  \\
    
    \midrule
    
    \textbf{\textit{Global Inference}} & & &   \\
    U-Net & 38.4  & -   & -  & 5507  \\
    ICNet  & 40.2  & - & - & 2557    \\
    PSPNet & 56.6 & - & - & 6289  \\
    DeepLabv3+ & 63.5 & -  & - & 3199   \\
    FCN-8s & 68.8  & 79.8 & 86.2 & 5227  \\
    BiseNetV1 & 53.0 & - & - & 1801  \\
    DANet & 53.8 & - & - & 6812  \\
    STDC & 70.3 & - & - & 2580  \\
    
    \toprule
    
    \textbf{UHR Model} &  &  &  &    \\ \midrule
    CascadePSP & 68.5 & 79.7 & 85.6 & 3236   \\
    PPN & 71.9 & - & - & 1193  \\
    PointRend & 71.8 & - & - & 1593  \\
    MagNet & 72.9 & - & - & 1559  \\
    MagNet-Fast & 71.8 & - & - & 1559  \\
    GLNet  & 71.6 & 83.2 & 88.0  & 1865 \\
    ISDNet & 73.3 & 84.0 & 88.7  & 1948  \\
    FCtL  & 73.5 & 83.8 & 88.3 & 3167  \\ 
    WSDNet & 74.1 & 85.2 & 89.1 & 1876 \\
    GPWFormer (Ours) & \textbf{75.8} & \textbf{85.4} & \textbf{89.9} & 2380  \\
     
    \bottomrule
    \end{tabular}}
    \caption{Comparison with state-of-the-arts on DeepGlobe \textit{test} set.}
    \label{sota_deepglobe}
\end{table}

\subsubsection{Comparisons of Downsampling Methods in WFormer}

We show the comparison of self-attention block with different downsampling methods in Table \ref{exp_downsample}. Classical self-attention with ordinary average pooling or transposed convolution shows much lower performance, since both of them are irreversible and lose much information during the downsampling operations, while wavelet transform is invertible thus all the information can be persevered. A self-attention block with transposed convolution achieves a high mIoU than with average pooling, but also introducing extra memory cost.

\begin{table}[ht]
    \centering
    \scalebox{1}{\begin{tabular}{l c c c c c}
      \toprule
      \textbf{Generic Model} & 
      \begin{tabular}[c]{@{}c@{}} mIoU \\ (\%)$\uparrow$ \end{tabular} & 
      \begin{tabular}[c]{@{}c@{}} F1 \\ (\%)$\uparrow$ \end{tabular} &  \begin{tabular}[c]{@{}c@{}} Acc \\ (\%)$\uparrow$ \end{tabular} & \begin{tabular}[c]{@{}c@{}} Mem \\ (M)$\downarrow$ \end{tabular}  
      \\ \midrule
    
    DeepLabv3+ & 55.9 & -  & - & 5122    \\
    FCN-8s & 69.1  & 81.7 & 93.6 & 2447 \\
    STDC & 72.4 & - & - & 7410  \\
    \toprule
    
    \textbf{UHR Model} &  &  &  &   \\ \midrule
    CascadePSP & 69.4 & 81.8 & 93.2 & 3236   \\
    GLNet  & 71.2 & - & -  & 2663 \\
    ISDNet & 74.2 & 84.9 & 95.6  & 4680  \\
    FCtL  & 73.7 & 84.1 & 94.6 & 4332  \\ 
    WSDNet & 75.2 & 86.0 & 96.0 & 4379 \\
    GPWFormer (Ours)  & \textbf{76.5} & \textbf{86.2} & \textbf{96.7} & 4710  \\
     
    \bottomrule
    \end{tabular}}
    \caption{Comparison with state-of-the-arts on Inria Aerial \textit{test} set.}
    \label{sota_inria}
\end{table}

\begin{table}[ht]
    \centering
    \scalebox{1}{\begin{tabular}{l c c c}
      \toprule
      \textbf{Generic Model} & mIoU (\%)$\uparrow$ & Mem (M)$\downarrow$ 
      
      \\ \midrule
    
    BiseNetV1 & 74.4 & 2147    \\
    BiseNetV2 & 75.8 & 1602    \\
    PSPNet    & 74.9 & 1584  \\ 
    
    DeepLabv3 & 76.7 & 1468  \\
    \toprule
    
    \textbf{UHR Model} &  &   \\ \midrule
    
    DenseCRF & 62.9 & 1575  \\
    DGF      & 63.3 & 1727  \\
    SegFix   & 65.8 & 2033  \\
    
    MagNet   & 67.6 & 2007  \\
    MagNet-Fast & 66.9 & 2007  \\
    
    ISDNet & 76.0 & 1510  \\
    
    GPWFormer (Ours) & \textbf{78.1} & 1897  \\
     
    \bottomrule
    \end{tabular}}
    \caption{Comparison with state-of-the-arts on Cityscapes \textit{test} set.}
    \label{sota_cityscapes}
\end{table}

\subsubsection{The Impact of Settings in Spatial Congruence}

The comparison of different congruence methods is shown in Table \ref{exp_congruence}. Here besides the Gaussian RBF kernel, we also implement some other metrics, including ``Mean" and ``Inner-Dot". The former indicates calculating inter-patch relation of two patches as distance between their mean feature, while the latter as the Inner-Product. Experiments shown that Gaussian RBF is more flexible and powerful in capturing the complex non-linear relationship between high-dimensional patch features.

\begin{table}[ht]
\centering
\scalebox{1}{\begin{tabular}{l|c|c}
\toprule
\multirow{1}{*}{\begin{tabular}[l]{@{}l@{}} \\ \textbf{Method} \\  \textbf{\ }\end{tabular}} & \multicolumn{1}{c|}{ISIC}                                                                                      & \multicolumn{1}{c}{CRAG}                                                                                       \\ 
                                                                       & mIoU (\%)   & mIoU (\%)   \\ \midrule
PSPNet                                                                 & 77.0                                                                                            & 88.6                                                                                          \\

DeepLabV3+                                                                & 70.5                                                                                             & 88.9                                                                                                \\
DANet                                                                  & 51.4                                                                                              & 82.3                                                                                               \\
GLNet                                                                  & 75.2                                                                                                & 85.9                                                                                               \\
GPWFormer (Ours)               
     & \textbf{80.7}                                                                                         & \textbf{89.9}                                                                                 \\ 
 \bottomrule
\end{tabular}}
\caption{Comparison with state-of-the-arts on CRAG and ISIC \textit{test} set.}
\label{sota_isic_crag}
\end{table}

\subsection{Comparison with State-of-the-Arts}

 In this section, we compare the proposed framework with existing state-of-the-art methods, including U-Net \cite{unet}, ICNet \cite{icnet}, PPN \cite{ppn}, PSPNet \cite{pspnet}, SegNet \cite{segnet}, DeepLabv3+  \cite{deeplabv3+}, FCN-8s \cite{fcn}, CascadePSP \cite{cascadepsp}, BiseNet \cite{bisenet}, PointRend \cite{pointrend}, DenseCRF \cite{densecrf}, DGF \cite{dgf}, DANet \cite{danet}, SegFix \cite{segfix}, MagNet \cite{magnet}, STDC \cite{stdc},  GLNet \cite{glnet}, FCtL \cite{fctl}, ISDNet  \cite{isdnet} and WSDNet \cite{urur}, on DeepGlobe, Inria Aerial, Cityscapes, ISIC and CRAG  datasets, in terms of mIOU (\%), F1 (\%), Accuracy (\%), Memory Cost (M). 

Some of these methods are specially designed for UHR images (denoted as ``UHR Model") and the others are not (denoted as ``Generic Model"). We show the results of ``Generic Model" on both ``Global Inference" and ``Local Inference". The former obtains the prediction with downsampled global image, and the latter obtains the prediction with local cropped patches sequentially and then merges their results by post-processing.

\subsubsection{DeepGlobe} 
As shown in Table \ref{sota_deepglobe}, we first compare GPWFormer with above-mentioned methods on DeepGlobe \textit{test} dataset. Due to the diversity of land cover types and the high density of annotations, this dataset is very challenging. The experiments show that GPWFormer outperforms all other methods on both mIoU, F1 and Accuracy. Specifically, we outperform GLNet, ISDNet and FCtL by large margins on mIoU respectively, directly showing the segmentation effectiveness and performance improvement. Besides, the categories in the dataset are often seriously unbalanced distributed, so we exploit the F1 score and Accuracy metrics to reflect the improvements and experiment results show the proposed method also achieves the highest scores among all the models. With such impressive performance, our methods is economic in the memory cost, attaining an excellent balance among accuracy and memory cost.

\subsubsection{Inria Aerial}  We also show the comparisons on Inria Aerial \textit{test} dataset in Table \ref{sota_inria}. This dataset is more challenging, since the number of pixels for each image reaches 25 million, which is around four times than DeepGlobe, and the foreground regions are also finer. Experiment results show that 
GPWFormer outperforms GLNet, ISDNet and FCtL by large margins again on mIOU respectively, with comparable memory cost.

\subsubsection{Cityscapes}
To further validate the generality of our method, we also show  the results on Cityscapes dataset, as shown in Table \ref{sota_cityscapes}. GPWFormer also outperforms all other methods on mIoU, with a bright results on memory cost.

\subsubsection{ISIC and CRAG}
The image resolution of ISIC is comparable to Inria Aerial, and CRAG is of lower image resolution than other datasets. Table \ref{sota_isic_crag} shows the experimental results. GPWFormer once achieves excellent performances.

\section{Conclusion}

In this paper, we focus on the ultra-high resolution image segmentation and develop a novel Guided Patch-Grouping Wavelet Transformer network.  
We firstly analyze the limitations of existing state-of-the-art methods, and the unique difficulties of ultra-high resolution image segmentation. 
With CNN-Transformer dual-branch, we propose the Wavelet Transformer with a Guided Patch-Grouping strategy to learn local details and long-range spatial dependencies simultaneously, while the CNN branch takes the downsampled input UHR image with wavelet transform to capture deep category-wise context. Moreover, the congruence constraint is also introduced to maintain the spatial consistency from CNN branch to Transformer branch. Our proposed framework achieves new state-of-the-art results on five benchmark datasets.

\section*{Acknowledgements}
This work was supported by the JKW Research Funds under Grant 20-163-14-LZ-001-004-01, Anhui Provincial Natural Science Foundation under Grant 2108085UD12,  National Key R\&D Program of China under Grant 2020AAA0103902, NSFC (No. 62176155), Shanghai Municipal Science and Technology Major Project, China (2021SHZDZX0102). We acknowledge the support of GPU cluster built by MCC Lab of Information Science and Technology Institution, USTC.

\bibliographystyle{named}
\bibliography{ijcai23}

\end{document}